\def\year{2020}\relax
\documentclass[letterpaper]{article} 
\usepackage{aaai20}  
\usepackage{times}  
\usepackage{helvet} 
\usepackage{courier}  
\usepackage[hyphens]{url}  
\usepackage{multirow}
\usepackage{multicol}
\usepackage{subfig}
\usepackage{graphicx} 
\usepackage{graphicx}  
\title{Learning to Order Sub-questions for Complex Question Answering }
\begin{document}
 \author{\\Yunan Zhang,\textsuperscript{1}
Xiang Cheng,\textsuperscript{1}
Yufeng Zhang, \textsuperscript{2}\\
Zihan Wang, \textsuperscript{1}
Zhengqi Fang,\textsuperscript{1}
Xiaoyan Wang, \textsuperscript{1}
Zhenya Huang, \textsuperscript{3}
Chengxiang Zhai\textsuperscript{1}\\
\textsuperscript{1}{University of Illinois at Urbana-Champaign}
\textsuperscript{2}{Bytedance Inc}\\
\textsuperscript{3}{University of Science and Technology of China}\\
\{yunanz2, xiangc2, zihanw2, xiaoyan5,zf4,czhai\}@illinois.com,\\
zhangyufeng.96@bytedance.com, huangzhy@mail.ustc.edu.cn}

\maketitle

\begin{abstract}
\begin{quote}
Answering complex questions involving multiple entities and relations is a challenging task. 
Logically, the answer to a complex question should be derived by decomposing the complex question into multiple simple sub-questions and then answering those sub-questions. 
Existing work has followed this strategy but has not attempted to optimize the order how those sub-questions are answered. As a result, the sub-questions are answered in an arbitrary order, leading to larger search space and higher risk of missing an answer.  
In this paper, we propose a novel reinforcement learning (RL) approach to answering complex questions that can learn a policy to dynamically decide which sub-question should be answered at each state of reasoning. We leverage the expected value-variance criterion to enable the learned policy to balance between the risk and utility of answering a sub-question. Experiment results show that the RL approach can substantially improve the optimality of ordering the sub-questions, leading to improved accuracy of question answering. The proposed method for learning to order sub-questions is general and can thus be potentially combined with many existing ideas for answering complex questions to enhance their performance. 
\end{quote}
\end{abstract}

\section{Introduction}
Real-world questions can be complex, involving multiple inter-related entities and relations, which we refer to as {\em complex questions}. For example, ``who writes Harry Potter" is a simple question that only involves a single entity and a relation, while ``Which city is the filming location of the book written by J.K.Rowling and held Olympics?" is a complex question, which consists of multiple entities and relations. How to automatically answer such complex questions is a significant scientific challenge because
it requires a system to capture the dependencies between different components of the questions and reason over them. As a result, most existing work on question answering can only handle simple questions, which have very limited application value.

Recently, some recent work has attempted to 
tackle such complex questions~\cite{DBLP:conf/naacl/TalmorB18,DBLP:journals/corr/IyyerYC16,DBLP:conf/acl/MinZZH19,DBLP:conf/acl/ZhangCXW19}, usually by decomposing a complex question into a sequence of simple questions and answering them based on a computation tree derived from the original question that can capture the dependency between sub-questions as shown in Figure 1. 

Logically, answering a complex question would require first decomposing the complex question into multiple sub-questions and then deriving a final answer to the question based on the answers to those sub-questions. In the example shown in Figure 1, we could first attempt to answer the sub-question 
``Which book was written by J.K.Rowling?" to obtain ``Harry Potter", which then allows us to obtain a second sub-question ``Which city is the filming location of Harry Potter?". Note that there is a dependency of the second sub-question on the first one since the second would not be well-defined if we did not have the answer to the first. Alternatively, however, we could have also attempted to first answer the sub-question ``Which city held Olympics?" before answering other sub-questions. 
In general, there may be many different orders one can use to answer those sub-questions; the more complex a question is, the more choices we would have in ordering the sub-questions. 

Does it matter which order to use for answering a complex question? The answer is yes. The dependency relation between sub-questions clearly requires an ordering of first answering the independent sub-question before answering a dependent sub-question. However, in addition to ordering based on dependency, there are still many different ways to order other sub-questions, and the order of answering those sub-questions generally has a significant impact on both the efficiency and the accuracy of question answering. 

To see why, consider how a human would order those sub-questions to answer the question shown in Figure 1. The dependency relationship is captured by the child-parent relation on the tree (a child question must be answered before its parent question can be answered). However, there are still different ways to order the leaf nodes and the intermediate nodes on different subtrees. When ordering those sub-questions, humans would consider not only the order of dependency but also the difficulty of the sub-questions and the information each can provide for answering subsequent sub-questions. Human would also tend to 
leverage connections between sub-questions. 

For example, in Figure 1, given the query ``Which city is the filming location of the book written by J.K.Rowling and held Olympics?", there are two entities (``Olympics" and ``J.K.Rowling") and three relations (``filming location", ``city of", and ``written by").  Humans can decompose it into several sub-questions and solve them in the following optimal order:  1) first answer the sub-question ``book written by J.K.Rowling" to obtain ``Harry Potter," 2) then find ``Harry Potter is filmed at which cities," and 3) finally, answer the sub-question ``Among these cities, which held Olympics?". This is an optimal order because in each step, the answer to a sub-question is essentially unique, leading to a very efficient inference chain to reach the answer without considering many other search paths along a sequence of sub-questions. 

Now let us consider the scenario of using a non-optimal order. For example, if we first attempt to resolve ``cities held Olympics," which is a very general question with many potential answers (since many cities have held Olympics), we would be ``lost" here in the sense it would be hard to know which of those cities we should further explore, or otherwise, we would have to simultaneously try all those cities, leading to an explosion of the search space. With limited resources in practice, we may not even be able to reach the answer at the end. Note that in this case, the order required by dependency relationship has not been violated. 

The example above shows clearly that the order does matter and optimization of the order is crucial to answering complex questions. Thus an important research question is: Can we use machine learning to learn an optimal order of those sub-questions so that a question answering system would be able to optimally process each sub-question to derive the final answer? Existing work has not addressed this research question and used only dependency-based partial ordering, which, as shown in the example above, is far from sufficient. 

 
\begin{figure}[t]
  \includegraphics[width=0.46\textwidth]{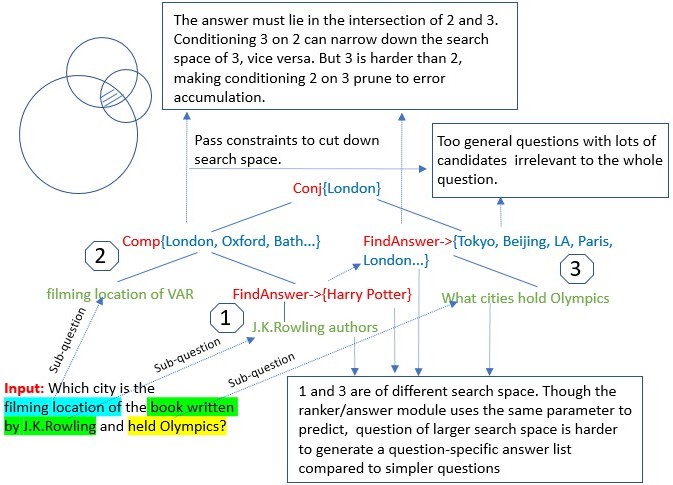}
  \caption{Issues of solving sub-questions without an optimized order.}
  \vskip-10pt
\end{figure}

In this paper, we address this problem by proposing a novel Reinforcement Learning (RL)-based approach to answering complex questions that can learn an optimal policy so as to 
intelligently choose
the sub-question most promising at each step in answering a complex question. 
The design of the RL framework is inspired by the following observations of how humans solve this problem: 
%
Humans would tend to pick up the answerable part of the complex queries first, use the partially solved parts as guidance to the rest of sub-questions, and finally narrow down the searching space to reach the answer. We are interested in studying how to 
use reinforcement learning (RL) to allow computers to learn this
problem-solving tactic automatically from training data.

Mastering such a problem-solving strategy requires the machines to solve three synergistic challenges: (1) understand the connections between sub-questions, (2) measure the utility and risk of questions, and (3) balance between utility and risk to dynamically decide which sub-question to solve at which time in order to efficiently and effectively solve the whole question. \par

To solve the first challenge, we map the sub-questions to sub-graphs in knowledge graph and encode them with graph convolution. This allows us to model the connections between sub-questions via multi-head attention~\cite{DBLP:conf/nips/VaswaniSPUJGKP17}. In this way, the model can leverage the graph structure of KB to learn connections and influences between sub-questions. For the second challenge, we model the utility of a sub-question via the correctness of the prediction and the distances between answer nodes and sub-graphs. The risk is measured in two aspects. One is the risk to enter error states. We give a negative reward to discourage this situation. The other is the uncertainty of the decision, which we capture by the variance of the returned reward.
To tackle the third challenge, we design the target function  with expected-value minus-variance criterion~\cite{DBLP:conf/icml/Heger94}. The balance between utility and risk is automatically optimized with policy gradient . Also, the reward depends on correctly answering the whole question. The agent would learn to order each sub-question safely to obtain the reward~\cite{DBLP:journals/corr/abs-1806-01830}. 

With these strategies, the RL framework can learn an optimized order which can help strategically decide what sub-question the model should attempt at certain states, with its risk and utility in mind. Specifically, the framework can learn to (1) avoid first answering risky uncertain sub-questions, (2) prioritize sub-questions of high utility and leverage their answer as guidance to help answer harder sub-questions, (3) balance between (1) and (2) since the most useful sub-question may not be the least risky one. \par

We evaluate the proposed RL framework on multiple data sets and the experimental results show that it can effectively learn to improve the ordering of sub-questions, leading to improvement of accuracy.  
The proposed RL framework and policy learning are general, and thus they can be combined with many existing techniques for answering complex questions to enhance their performance. 


\begin{figure*}
  \includegraphics[width=\textwidth]{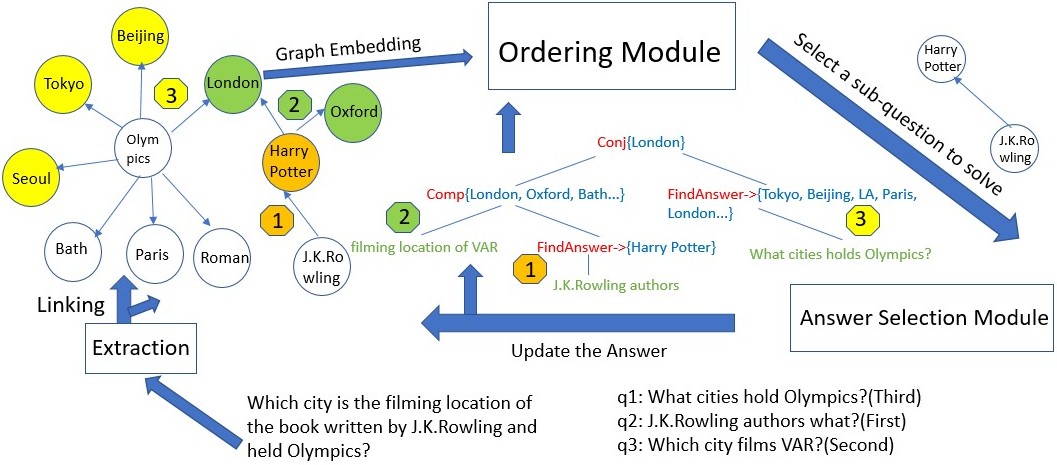}
  \caption{General pipeline of our model. Every module can be easily replaced with more sophisticated ones.}
  \vskip-10pt
\end{figure*}
\section{Related Work}

Our work is the first to study how to learn an optimal order of answering sub-questions for answering complex questions.  
The definition of complex question answering varies, but it generally refers to answering questions that require thinking processes more than keyword retrieval and extraction, like multi-hop reasoning and external knowledge acquisition. In this paper, we focus on studying questions that involve multiple entities and relations, which is close to the multi-hop reasoning setting. Our definition is similar to that in Complex Web Questions~\cite{DBLP:conf/naacl/TalmorB18}, where a complex question can be decomposed into several simple questions, whose dependency can be represented by a tree.\par
Question decomposition is one of the mainstream methods to answering complex questions~\cite{DBLP:conf/naacl/TalmorB18,DBLP:journals/corr/IyyerYC16,DBLP:conf/acl/MinZZH19,DBLP:conf/acl/ZhangCXW19}. ~\cite{DBLP:conf/naacl/TalmorB18} and~\cite{DBLP:conf/acl/MinZZH19} use Pointer Network~\cite{DBLP:conf/nips/VinyalsFJ15} to generate split points that separate the original complex question into several spans, where each span is a simple question. Dependency among sub-questions and logic operations like conjunction are jointly decided via seq2tree.
We adopt a similar divide-and-conquer process but focus mainly on the order of conquering sub-questions. Also, they solve each sub-question independently, without information from the other parallel sub-questions and the whole question. Our work shares some spirits of ~\cite{DBLP:journals/corr/IyyerYC16} in that we both leverage the idea that simpler questions are easier to predict for neural networks. But their method does not optimize the order. They split the complex question into sequential simple question via crowdsourcing and then solve each sub-question to get the final answer via semantic parsing. The work focuses mainly on semantic matching and coreference resolution.
~\cite{DBLP:conf/acl/ZhangCXW19} splits the question in the embedding space instead of conducting token level splitting. The split questions are then translated into structured SPARQL query via semantic parsing. However, this method also suffers from error accumulation both in the splitting stage and the semantic parsing stage, which could be alleviated via a more intelligent planning on the decoding process as we attempt to achieve in our work.

Our way of optimizing order of sub-questions is closely related to risk sensitive decision making. Managing the risk and value of a decision is broadly studied in the reinforcement learning community~\cite{DBLP:conf/icml/Heger94,DBLP:conf/nips/NeuneierM98,DBLP:conf/emnlp/LinSX18,DBLP:journals/jair/GeibelW05}. However, the idea is rarely explored in NLP applications, like complex question answering and semantic parsing which also make a sequence of decisions. The utility is mainly related to the value concept in RL. The risk can be categorized into two kinds. One is the inherent uncertainty of nature. That is, even with the optimal policy, the stochastic nature of the environment could still produce unwanted outcomes. This kind of risk is usually measured via the variance of the returned reward. The other kind of risk is defined as the probability of visiting the error state. This risk is usually modeled via a negative reward to penalize entering an easy-to-fail state. We apply this idea in answering complex questions via optimizing the order of answer sub-questions. With the learned order, the agent avoids risky states, namely answering harder questions too early, but work on risky ones after more information is gained, which could hopefully lower the risk.

Our work is closest to the work 
~\cite{DBLP:conf/emnlp/XiongHW17,DBLP:conf/iclr/DasDZVDKSM18,DBLP:conf/naacl/GodinKM19}, which uses reinforcement learning to answer questions on knowledge graph. These works mainly focus on handling the incompleteness of KG via multi-hop reasoning. The multi-hop reasoning is used to find a path formed by multiple relations that is equivalent to a missing link in KG. For example, given the question`` Who is Tom's grandfather". On KG, Tom and his grandfather may not be connected via the relation grandfather, namely, the link is missing. But the relation father exists between Tom and Tom's father, as well as the father of Tom's father. Following the father's father path, we actually find an alternative of the missing grandfather link. In this way, the question can still be answered via the inducted path. Our work makes novel contributions in learning to optimize the order of answering sub-questions via RL and using knowledge graph structure to capture connections among sub-questions, but our answering module is similar to theirs. It is worth noting that in this work, we focused on the learning of an optimal order, thus we used a relatively simple answering module, which can be easily upgraded to more sophisticated answering module without affecting the learning mechanism of the ordering. 
\section{Problem Definition}
In this section, we formally define our problem.  We define a complex question as a question containing multiple entities and relations while a simple question as one with only one entity and one relation. A knowledge graph can be represented as $\mathcal{G = \{\mathcal{E,R} \}}$, where $\mathcal{E}$ is the set of entities and $\mathcal{R}$ is the set of relations. A KB fact, or triple, is stored as, ($e_1$, r, $e_2$) $\in \mathcal{G}$, where $e_1$ is the source entity, $e_2$ is the target entity, r is the relation connecting $e_1$ and $e_2$. 

In general, a natural language question can be parsed to generate a structured question representation, including a query set $\{e_1,..., e_n, r_1,..,r_m\}$, and a computation tree, where $e_i$ and $r_j$ are an entity and relation mentioned in the original complex question, respectively. The computation tree is defined in a similar way to that in Complex Web Questions, where leaf nodes are strings corresponding to sub-questions, and inner nodes are functions applied to strings, e.g, answering function, conjunction operations, and comparison operations. Note that there is no guarantee that the answering function is perfect, thus causing error accumulation when earlier answers are wrong. Also, the sibling nodes (i.e., results of each sub-question) are actually not independent and can provide useful information to other sub-questions. 

Our goal is to (1) optimize the order of answering to alleviate the first issue, and also (2) help  strategically collect information according to utility and share the newly gained information among sub-questions to assist answering harder sub-questions.\par

We chose this form of input because it is a common output that can be realistically generated by an NLP parser based on the free text question and would allow us to focus on studying how the ordering matters and how to optimize the order.  

Note that $\forall i, e_i \in \mathcal{E}$, but $\exists i, r_j \notin \mathcal{R}$ due the incompleteness of KB~\cite{DBLP:conf/emnlp/XiongHW17,DBLP:conf/emnlp/LinSX18}. 
Each sub-question is defined based on an entity $e_i$ in the original question and a set of relations $r_j$ assigned to $e_i$, namely $R_i$. Answering a sub-question is completing an inference chain starting from $e_i$, $(e_i,r_{i1},e_{o1}), (e_{o1}, r_{i2}, e_{o2})...(e_{o|R_i|-1}, r_{i|R_i|}, e_{o|R_i|})$, where $e_{ot}$ is the returned answer from previous step. We allow sub-questions to have branches, namely the newly picked r may not be assigned to the last inferred entity but can be assigned to any inferred entities. 
Taking an intersection ensures that the answer has an inference link with every entity in the
original question. 

\section{Reinforcement Learning Formulation}
Given a complex query in the form $e_1,..., e_n, r_1,..,r_m$. We first link each entity to the given KG. Each entity is used as the initial point of each sub-graph.  The question answering process can be viewed as a cooperated search among the sub-graphs over $\mathcal{G}$, which is formulated as a Markov Decision Process (MDP). The initial state of the MDP consists of a set of sub-graphs, each corresponding to one distinct entity in the original question, and the relations in the original question. The agent can take an action to extend any of the subgraphs with a relation connected with the subgraph (according to KG) to extend the subgraph, causing the state to be transitioned to another state which differs from the current state mainly in the subgraph chosen to expand. However, as the action also involves the use of a relation $r$ to expand the subgraph, the relation part of the state is also updated to reflect the fact that relation $r$ has been ``consumed" by reducing its weight.   
We now describe the proposed RL framework more formally, covering Action Space, States, Reward Design, Target Function, and Policy Network. 

\subsection{Action Space} 
The action space can be written as:\par $A_t = \{(e_s,\hat{r_t}, e_t)|e_s \in \{g_{ti}\}_{i=1}^n\, (e_s,\hat{r_t}, e_t)\in \mathcal{G}\}$,\\ \\namely choosing a triple whose source entity $e_s$ is in the current state (sub-graphs). We have a self-loop action for every node to indicate the answer.
$\hat{r_t},e_t$is outgoing edge of $e_s$ and the reached entity. Transition can be represented as a probability matrix:\par 
$P(s_{t+1} = s'| s_{t}=s, a_t = a)$. \\We describe how our proposed solution to (1)capturing connections between sub-questions and (2)order sub-question solving sequence, in the state representation and reward design section.\\

\subsection{States}
\begin{figure}[t]
  \includegraphics[width=0.5\textwidth]{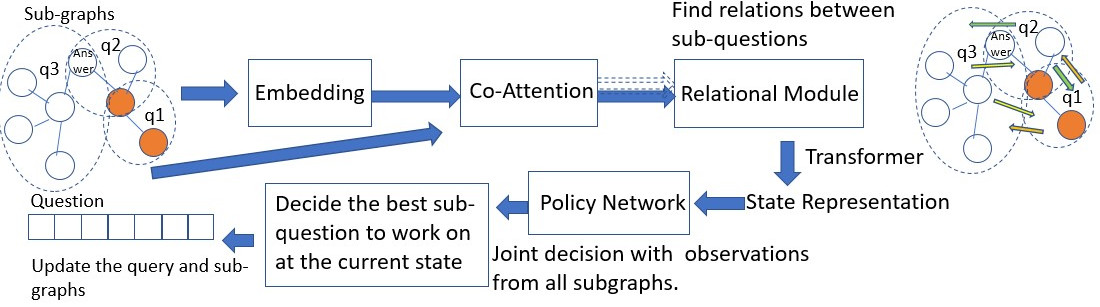}
  \caption{The pipeline of the ordering module}
\end{figure}
To perform the cooperative search, we want the agent to not only know the local observation of the sub-question, but also the observation from other sub-questions. So we fuse the observations from all sub-questions according to relation between them to enable the agent make the best plan based on the  joint information. The state contains the aggregation of all the sub-graphs and the query state:\par
$s_t = [(h'_{t1}; h'_{t2}...;h'_{tn}), (R_t)]$,\\
$h'_{ti}$ represents the sub-graph i at time step t, which is initialized as the entity node mentioned in the question. $R_t$ represents the state of the original query and is initialized as:\par $R_0 = [r_1,..,r_m]$. \\We discuss how to update them in the following sections.\par
\textbf{Sub-graph Representation:}
For each sub-graph, we encode their information as a weighted sum of node embedding conditioned on the query state. The weight of nodes reflects how likely the node would be relevant to the current step of reasoning, given the query state $R_t$. Assume a sub-graph $g_{ti}$ has k nodes, \par $g_{ti} = e_{t1}...e_{tk}$, \\at time step t. To capture the semantic and graph structure information of the entities, we use ConvE~\cite{DBLP:conf/aaai/DettmersMS018} to encode them. Then at each time step t, we compute an affinity matrix of the sub-graph and query to help the model get the importance of each node w.r.t the query state:\par
$L = g_{ti} * R_t$, where $g_{ti} = e_{t1}...e_{tk}$\par
$A^{R_t} = softmax(L) \in R^{k*m}$,\par
$h_{ti} = g_{ti} * A^{R_t} \in R^{l * k}$.\\
Here, we get the representation of sub-graph i as $h_{ti} = g_{ti} * A^{R_t}$, where $A^{R_t}$ is the attention weights across the sub-graph for each relation in $R_t$. $h_{ti}$ is the attention context of $g_{ti}$ in light of $R_t$. In this way, we get the representation of an individual sub-graph. We then discuss how to aggregate information from multiple sub-graphs based on their interactions in next section.
We then discuss how to update the query as the reasoning process goes on.\par
\textbf{Interaction between Sub-graphs:}
 The solving of each sub-questions are correlated to each other as the knowledge about them are related someway on the KG. The knowledge triple stored in KG is related to each other via graph structure and semantic clues. Understanding the clues between knowledge and their relevance to the given question can make reasoning more efficacious. Human brains can capture the connections via commonsense knowledge on sub-questions and connect the dots efficiently with logic. However, it's challenging for machines to induce relatedness between concepts. In our model, sub-questions are mapped to sub-graphs of KB. But different sub-graphs in KB may not be in the neighborhood of each other. This requires our system to model non-local interactions. We model this interaction as a message passing process. The relation between the ith and jth sub-graph can be written as:\par
$\alpha_{ij}^m = \frac{exp(\tau * W_q^m h_i * (W_k^m h_j)^T)}{\sum_{e \in \varepsilon_i}exp(\tau * W_q^m hi * (W_k^m h_e)^T)}$,\par
$h_i^{'} = \sigma(Concat[\sum_{j \in \varepsilon_i}\alpha_{ij}^m W_v^m h_j]), \forall m \in M$.\\
Here, we adopt self-attention to capture the relations between different pairs of sub-graphs. m is the number of attention heads we use. $W_k, W_v, W_q$ are parameters of key, value, and query that can be learned. In this way, we aggregate the information from the sub-graphs so that each sub-graph can grow itself with the knowledge of the other sub-graphs. In a pathfinding scenario, the knowledge from shared observation could help each agent find more efficient ``short cut" towards the answer since it has multiple viewpoints from teammates now. The decision made is then conditioned on the whole team. After the decision is made, we need to update the query and graphs with the newly gained information. \par
\textbf{Query Reduction:}\\  When humans conduct compositional reasoning, it's intuitional to substitute the used entity and relation pair with the newly inferred entity. For example, in the ``Which city in UK is the filming location of the book written by J.K.Rowling?", people won't be interested in ``J.K.Rowling" after they've inferred ``Harry Potter". With this in intuition, each time a sub-graph chooses an outgoing edge, $\hat{r_t}$ from its neighborhood, we update the sub-graph as well the query state$R_t$. The sub-graph update itself by adding the newly arrived node while $R_t$ updates by decreasing the weights of relations relevant to $\hat{r_t}$:\par
$R_t = R_{t-1} - \gamma * \hat{r_t}$.\\
$\gamma$ is a matrix computing the similarity between $\hat{r_t}$ and each relation in $R_{t-1}$. Since the sub-graph representation also depends on $R_t$, we also implicitly lower the weights of the nodes used previously by lowering the weights of relations related to them. In this way, the model would be less vulnerable to disturbance from the finished inference.\par
\textbf{Final Representation:}
Then we can write state at time step t as: $s_t = [(h'_{t1}; h'_{t2}...;h'_{tn}), (R_t)]$, where $(h'_{t1}; h'_{t2};...;h'_{tn})$ are the concatenation of the sub-graphs. $R_t$ is the query states. We also record $Y_t = (e_{t1}, e_{t2}...e_{tn})$which are the last hit node for each sub-graph till time step t.\par
\subsection{Reward Design}
We design the reward with two main principles. One principle is to encourage the agent to pick sub-questions with high utility, namely to what level does its answer help makes the prediction of the whole question correct. The other principle is avoiding the sub-questions that have a high risk of attempting at the given state,  namely how likely the prediction made for a sub-question is wrong at the given state. The idea is drawn from the expected value-variance criterion which states that a good decision sequence should balance well between return value and the risk of getting the value.\par

{\bf Risk:}
We define the risk of decisions in two aspects. One is the uncertainty of taking a sub-question, namely the variance of the return, the other being the likelihood of entering error states. The first aspect can be calculated via the variance of the expected reward. The second aspect is learned via assign a negative reward, -1, to the wrong prediction. Policy gradient would then optimize the decision sequence automatically given the target function.\par  
In experimental analysis, we compute the risk of taking a sub-question as the probability of answering the sub-question wrongly at a given state:\\
$P(answer=negative|state=s, q=i) = P(a=negative, q=i | state = s)/P(q = i|state = s)$.\\
Here $P(a=negative, q=i | state = s)$ and $P(q = i|state = s)$ can be derived from the output of neural network. 

{\bf Utility:}
We measure the utility of a sub-question in two ways. One is how it can help other sub-questions to cut down search space. The other is whether it helps answer the whole question. The later is easy, which is completing the task successfully on termination. We assign +1 reward for correctly predict the answer for the whole question. The former can be modeled as helping more sub-questions reach the answer. So we encourage the case where more sub-graphs reach the answer when the search terminates. Assume there are z sub-graphs grow to the correct entity, the agent receives $R(s_T) = z * 1 + \lambda * y$, where y is the number of sub-graphs interact correctly. To reduce variance, we subtract an average cumulative discounted reward scaled by a hyperparameter.
\subsection{Target Function}
Combined with the later parts, we now write our target function as:\par$J(\theta) = E_{q \in Q}[E_{\tau \sim \pi_\theta}[R(s_T|q)] - k* var(R(s_T|q))]$,\\ where k is a small positive number, q is a query in the query set Q, $\theta$ is the parameters of the policy network. We'll discuss the policy network and optimization in next section.
\subsection{Policy Network}
We encode the search history $p_t$ as state action pairs via LSTM~\cite{DBLP:conf/nips/HochreiterS96}:\par $p_0 = LSTM(0,[R_0, H_0])$, $p_t = LSTM(p_{t-1}, a_{t-1})$.\\
The policy network can be written as:\par $\pi_{\theta}(a_t|s_t)=\sigma (At \times W_2 ReLU(W_1[[h'_{t1};...;h'{tn}];p_t;R_t]))$,\\ where $\sigma$ is the softmax function. It can be trained via maximizing reward over all queries Q with the target function mentioned above, which can be optimized via REINFORCE~\cite{DBLP:journals/ml/Williams92}. Parameters of the network,$\theta$, can be updated with stochastic gradient:\par $\nabla J(\theta) = \nabla_\theta \sum^T_{t=1}(R(s_T|q)-k*var(R(s_t|q)))log(\pi_\theta(a_t|s_t))$.
\section{Experiments}
\subsection{Datasets}
We test our ideas on four datasets. Complex Web Question ~\cite{DBLP:conf/naacl/TalmorB18} and three newly generated datasets constructed from Countries~\cite{DBLP:conf/icml/TrouillonWRGB16}, FB15k~\cite{DBLP:conf/emnlp/ToutanovaCPPCG15}, which is based on the Freebase\cite{DBLP:conf/sigmod/BollackerEPST08}
knowledge graph  and WC-14~\cite{DBLP:journals/corr/ZhangWT16}. The newly generated datasets are created in a similar way to that of Complex Web Questions. The difference is that we only generated conjunction questions.Question are shuffled and divided into training set and test set. The datasets statistics are summarized in Table 1. We report hit@1, hit@3, hit@10 for Countries, FB15k and WC-14 to test our model performance. We report accuracy for Complex Web Question. For Complex Web Questions, we show naive order-changing and information sharing can bring significant improvement. \\
\\
\begin{table}
\begin{tabular}{ |p{3.5cm}|p{1.5cm}|p{1.5cm}|  }
 \hline
 Dataset & Train & Test \\
 \hline
 Countries & 434 & 145\\
 \hline
 WC-C   & 803    & 327\\
 \hline
 FB15k & 15000 & 5001\\
 \hline
 Complex Web Questions & 27734 & 3475\\
 \hline
\end{tabular}
\caption{Datasets Summary}
\end{table}
\subsection{Testify the Order Matters}
We testify our hypothesis of (1)order matters, (2)sub-questions are dependent, on Complex Web Question. We made two changes to their pipeline. The first is to prioritize sub-questions whose output has higher precision. The other is to share information among sub-questions that are input of a conjunction. When a sub-question finishes computation, it will pass its result, which is a rank to its siblings. The sub-question receiving message will rank the intersection elements higher in its list. This simple re-ordering and information passing can bring around 2\% on precision@1, which proves the effect of better ordering and information sharing.
\begin{table}
\begin{tabular}{ |p{3cm}|p{1.2cm}|p{1.2cm}|p{1.2cm}| }
 \hline
 ~ & Countries & WC-C & FB15k\\
 \hline
 Attention Hidden Layers & 8 & 16 & 16\\
 \hline
 LSTM Hidden Layers   & 32    & 32 & 64\\
 \hline
\end{tabular}
\caption{Parameter Settings}
\vskip-10pt
\end{table}

\begin{table}
\begin{tabular}{ |p{1.9cm}|p{1.2cm}|p{1.2cm}|p{1.2cm}| }
 \hline
 ~ & Countries & WC-C & FB15k\\
 \hline
 Time(hours) & 1 & $<$1 & 15\\
 \hline
 Iterations   & 800    & 10 & 30\\
 \hline

\end{tabular}
\caption{Training time and iterations}
\end{table}
\begin{table*}[t]
\begin{tabular}{|l|l|l|l|l|l|l|l|l|l|}
\hline
\multirow{0}{1pt}{Models} & \multicolumn{3}{l|}{Countries} & \multicolumn{3}{l|}{WC-C} & \multicolumn{3}{l|}{FB-15k} \\ \cline{2-10} 
                        & Hits@1   & Hits@3   & Hits@10  & Hits@1 & Hits@3 & Hits@10 & Hits@1  & Hits@3  & Hits@10 \\ \hline
Ours                    & \textbf{0.551}    & \textbf{0.621}    & \textbf{0.717}    & \textbf{0.566}  & \textbf{0.608}  & \textbf{0.663}   & \textbf{0.210}   & \textbf{0.255}   & \textbf{0.389}   \\ \hline
\cite{DBLP:conf/iclr/DasDZVDKSM18}      & 0.481    & 0.523    & 0.631    & 0.421  & 0.523  & 0.619   & 0.184   & 0.237   & 0.267   \\ \hline
(Lin et al. 2018)      & 0.495    & 0.561    & 0.667    & 0.439  & 0.542  & 0.639   & 0.190   & 0.241   & 0.284   \\ \hline
\end{tabular}
\caption{Overall Performance}
\end{table*}
\subsection{Network Training}
\textbf{Training Setup.}
Our model constitutes three parts, a state encoding network, a two-layer LSTM for tracking the answered sub-questions and a policy network for selecting actions.  The state encoding network takes the current sub-graph representation vectors and encodes them with multi-head attention. The output will be concatenated with the output of the LSTM and feed into a two-layer policy network with a softmax layer at the end to output distribution of the actions.  The dimension of layers is different for different datasets. We trained our model on GTX 1080Ti. Table 3 shows the training time and iterations for each dataset.\\
\textbf{Initialization.}
For Countries, we use one-hot to encode the relation. For all other datasets, we used GloVe 300-dimension pre-trained word embeddings. Nodes are embedded with ConvE in all datasets.\\
\textbf{Optimization.}
We trained our model using policy gradient with Adam Optimizer. We also adopt the additive control variate baseline described in \cite{DBLP:conf/iclr/DasDZVDKSM18}.
\subsection{Performance}

In this section, we show the overall performance of our model on the Countries, FB15k, and WC-14. One thing to notice is that our proposed model is mainly intended to learn an optimized order instead of designing a new sophisticated question answering model. As shown in Figure 2, the ordering module and answering module can be conceptually separated. To make a fair comparison, we choose models that use similar answering modules to us. We use MINERVA~\cite{DBLP:conf/iclr/DasDZVDKSM18} and reward shaping ~\cite{DBLP:conf/emnlp/LinSX18} as baselines of our model. During experiments, the baselines have access to the computation tree derived from the question, namely, they won't attempt sub-question whose dependency is not satisfied. Under this constraint, their computation order is random. We summarized the performance in Table 4. The best score for each dataset is highlighted.\par

\section{Analysis}
In this section, we analyze why the performance improves with a better order and why our model learns a good order.\\
\\
\textbf{Why order matters.}
We analyze the error rate of our models and baselines made at each step of reasoning. As shown in Figure 4, the error rate of our model increases with the step, indicating the model chooses less risky ones at start. Also, the error rate of our model is lower than the baselines at the first two step. For the last step, our model has a slightly higher error rate. It's probably due to the fact that during implementation, our model doesn't fully separate the ordering and answering module. The network has to both learn the select a promising sub-question and learn to solve the sub-question with the same set of parameters. Therefore, even the model attempts the right question, it could still fail due to the fact that the parameters may not be optimized for the answering task.\\

\begin{figure}[ht!]
    \centering
    \subfloat{%
        \includegraphics[width=1.0\linewidth]{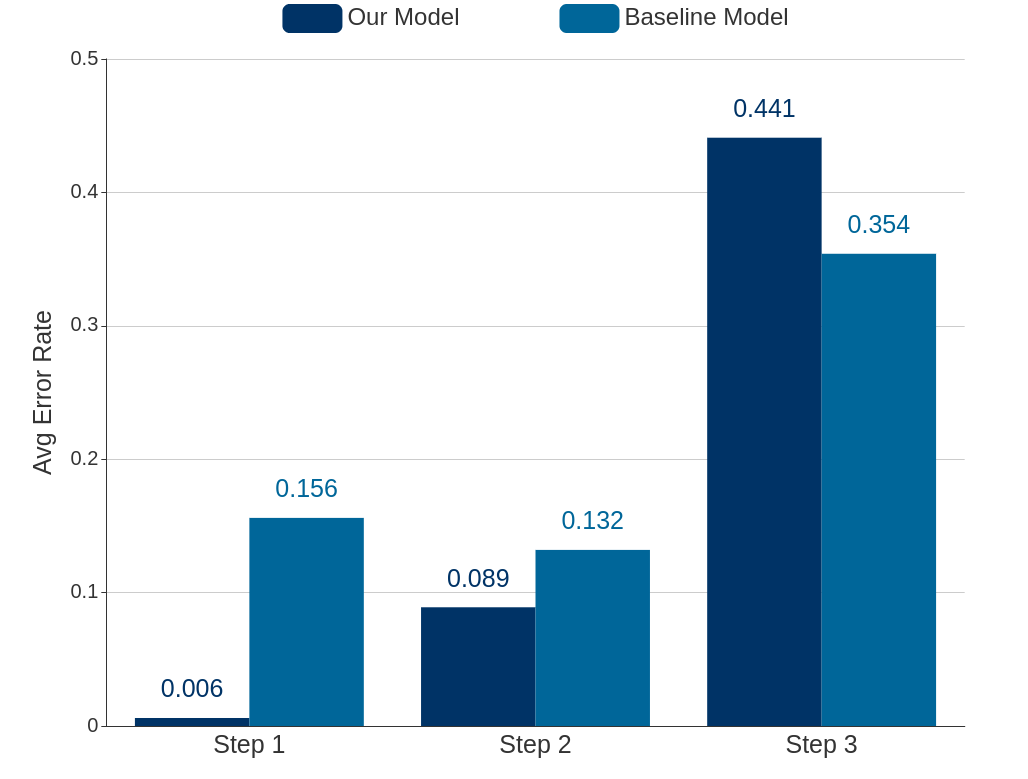}%
        }%
       \caption{Error rate at each step.}
\end{figure} 
\begin{figure}[ht!]
    \centering
    \subfloat{%
        \includegraphics[width=0.9\linewidth]{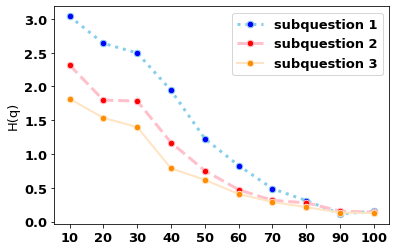}%
        }%
       \caption{Error rate at each step}
    \subfloat{%
        \includegraphics[width=0.9\linewidth]{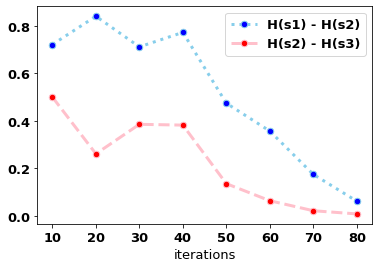}%
        }%
        \caption{Entropy loss of solving each sub-question through iterations}
        
    \subfloat{%
        \includegraphics[width=0.9\linewidth]{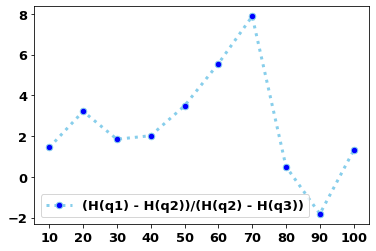}%
        }%
        \caption{Entropy ratio through iterations}
    \vskip-10pt
\end{figure}
Let $H(s_i)$ denotes the the entropy of the answer distribution output by the policy network at a given step i. The first graph shows $H(q_i)$ for each step of the reasoning during the training time. Notice that H(s) are evaluated on the test set. We can see from graph 1 that H(s1) > H(s2) > H(s3) which shows that answering a question at a step helps reducing the uncertainty for answering the next question. We can also see from graph 2 and graph 3 that $(H(s_1) - H(s_2))/(H(s_2) - H(s_3))$ are increasing. This shows that our model are learned to answer the question that has a larger utility first.\\
\textbf{Why our model learns a good order.} We analyze the relation between the sub-question selection and their corresponding utility and risk at each step. As we can see in Figure 4 and 5, the most risky sub-question becomes less risky after the completion of the other two sub-questions. Also, the model chooses the least risky sub-question first instead of the choosing sub-questions with a higher chance to fail. The utility can be seen from the entropy decreases ratio. Our model answers the question that cause larger decrease in entropy, indicating better clarification effects.
\begin{figure}[ht!]
    \centering
    \subfloat{%
        \includegraphics[width=0.9\linewidth]{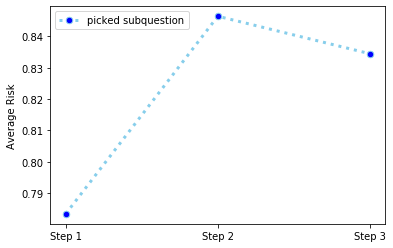}%
        }%
        \caption{Risk of choice at each step through iterations}
    \subfloat{%
        \includegraphics[width=0.9\linewidth]{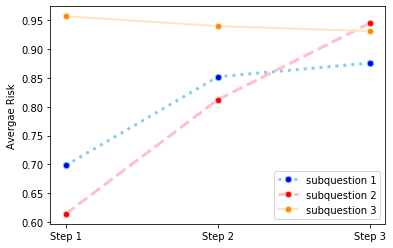}%
        }%
        \caption{Risk of each sub-question through iterations}
        \vskip-10pt
\end{figure}
\section{Conclusions and Future Work}
In this work, we studied how to optimize the order of answering a sequence of sub-questions decomposed from a complex question and proposed a novel RL framework to learn the optimized order of answering sub-questions. To achieve this goal, we propose a novel state representation that captures interactions between sub-questions via leveraging graph structure of KG and multi-head attention. We also introduce the concepts of risk and utility and incorporate them into the target function. Experimental results show that the proposed RL framework is effective for improving the order of answering sub-questions, leading to improvement of accuracy. 

Though we have shown our model can learn ordering that helps complex question answering, there is still much room to further extend or generalize the method, opening up many interesting directions for future research. First, our model is currently limited to knowledge graph settings and has restrictions on the input form; the requirement can be relaxed via incorporating more complex extraction modules and answering modules. Second, the learning mechanism for optimizing the order is general
and can not only be combined with other techniques for answering complex questions, but also be further generalized for optimizing machine reading comprehension and semantic parsing. Indeed, semantic parsing usually adopts seq2tree, which generates code from root to leaf. The children predictions are conditioned to their parents' predictions. This generation procedure is also prone to error accumulation from root level predictions. The proposed RL framework can potentially be used to optimize the decoding order and improve performance across many models. 
\section{Acknowledgments}
We thank the anonymous reviewers for providing insightful critics for this paper. We thank Suhansanu Kumar for giving valuable comments on this work. The first author's effort is dedicated to Ayanami Rei.
\bibliography{aaai20}
\bibliographystyle{aaai20}
\end{document}